%% file: emnlp2023.tex
\pdfoutput=1

\documentclass[11pt]{article}

\usepackage{EMNLP2023}

\usepackage{times}
\usepackage{latexsym}

\usepackage[T1]{fontenc}

\usepackage[utf8]{inputenc}

\usepackage{microtype}

\usepackage{inconsolata}

\usepackage{graphicx}
\usepackage{subcaption}
\usepackage{xspace}
\usepackage{tabularx}
\usepackage{booktabs}
\usepackage{amsmath}
\usepackage{amssymb}

\def\sys{CmdCaliper\xspace}
\def\dataset{CyPHER\xspace}

\title{CmdCaliper: A Semantic-Aware Command-Line Embedding Model and Dataset for Security Research}

\author{
 \textbf{Sian-Yao Huang\textsuperscript{1}},
 \textbf{Cheng-Lin Yang\textsuperscript{1}},
 \textbf{Che-Yu Lin\textsuperscript{1}},
 \textbf{Chun-Ying Huang\textsuperscript{2}}
\\
\\
 \textsuperscript{1}CyCraft AI Lab, Taiwan\\
 \textsuperscript{2}Department of Computer Science, National Yang Ming Chiao Tung University, Taiwan
\\
 \normalsize{
   \{eric.huang,cl.yang,jerry.lin\}@cycraft.com, chuang@cs.nycu.edu.tw
 }
}

\begin{document}

\newcommand{\eric}[1]{\textcolor{blue}{#1}}
\maketitle

\input{emnlp2023-latex/paper_content/abstract}
\input{emnlp2023-latex/paper_content/introduction}
\input{emnlp2023-latex/paper_content/related}
\input{emnlp2023-latex/paper_content/approach_dataset}
\input{emnlp2023-latex/paper_content/approach_model}
\input{emnlp2023-latex/paper_content/eval_dataset}
\input{emnlp2023-latex/paper_content/eval_model}
\input{emnlp2023-latex/paper_content/conclusion}

\input{emnlp2023-latex/paper_content/limitation}
\input{emnlp2023-latex/paper_content/acknowledgment}

\bibliography{anthology,custom}
\bibliographystyle{acl_natbib}

\appendix

\input{emnlp2023-latex/paper_content/appendix}

\end{document}

%% file: emnlp2023-latex/paper_content/abstract.tex
\begin{abstract}

This research addresses command-line embedding in cybersecurity, a field obstructed by the lack of comprehensive datasets due to privacy and regulation concerns. We propose the first dataset of similar command lines, named \dataset\footnote{\dataset: CyCraft's Paired Command-Lines Harnessed  for Embedding Research}, for training and unbiased evaluation. The training set is generated using a set of large language models (LLMs) comprising 28,520 similar command-line pairs. Our testing dataset consists of 2,807 similar command-line pairs sourced from authentic command-line data.


In addition, we propose a command-line embedding model named \sys, enabling the computation of semantic similarity with command lines. Performance evaluations demonstrate that the smallest version of \sys (30 million parameters) suppresses state-of-the-art (SOTA) sentence embedding models with ten times more parameters across various tasks (e.g.,  malicious command-line detection and similar command-line retrieval).

Our study explores the feasibility of data generation using LLMs in the cybersecurity domain. Furthermore, we release our proposed command-line dataset, embedding models' weights and all program codes to the public. This advancement paves the way for more effective command-line embedding for future researchers.
\end{abstract}


%% file: emnlp2023-latex/paper_content/introduction.tex
\section{Introduction}
\label{sec:intro}

\begin{figure}[t]
\centering
\begin{center}
   \includegraphics[width=0.9\linewidth]{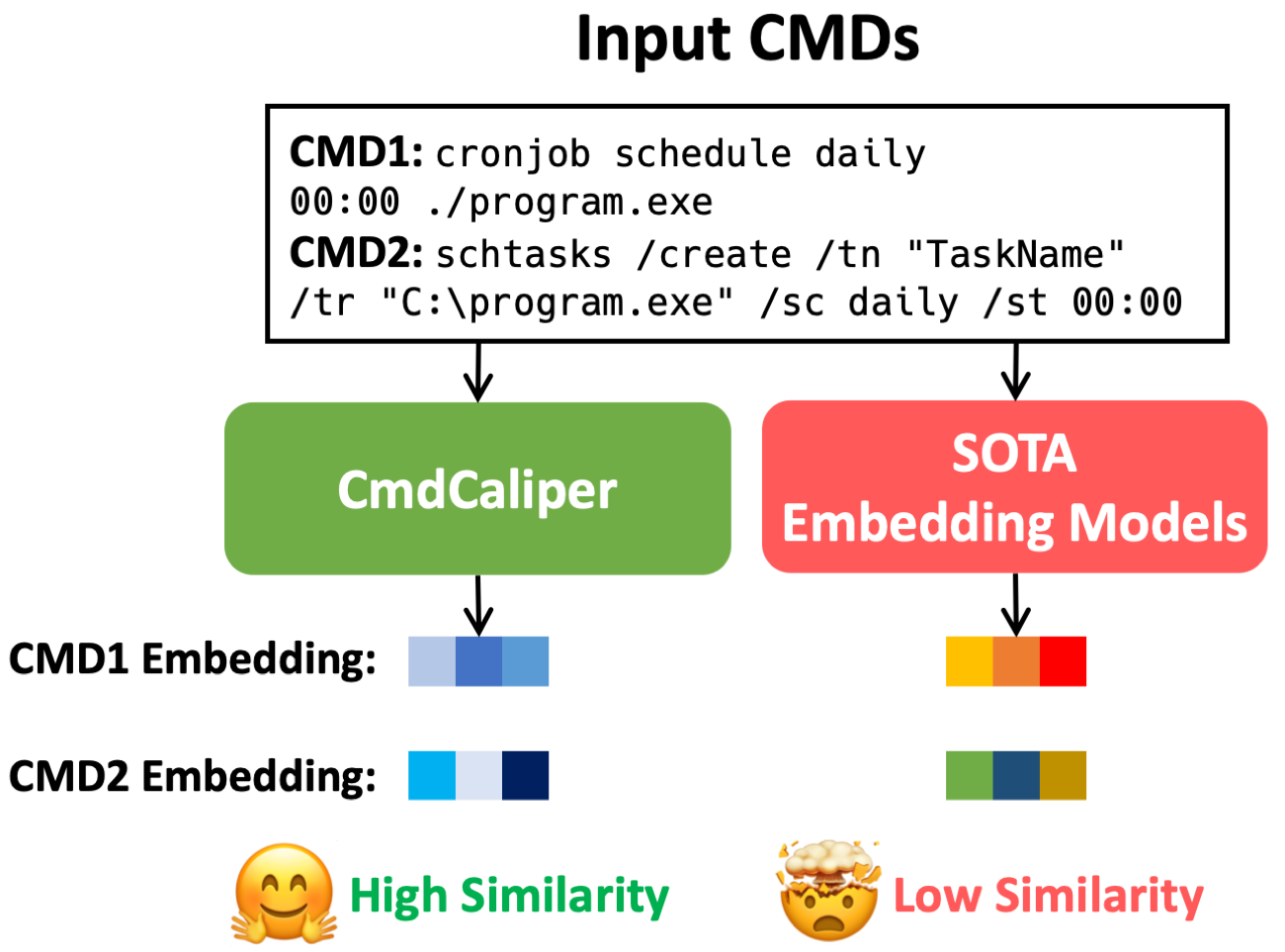}
\end{center}
   \caption{After fine-tuning our proposed similar command-line pair dataset, \dataset, our proposed command-line embedding model, \sys, can effectively embed command lines based on their semantics rather than solely on appearance.}
\label{fig:cover}
\end{figure}

Sentence embeddings, which map diverse sentences into a unified semantic feature space, are critical for various NLP applications such as classifier training, visualization~\cite{tsne}, and retrieval-augmented generation (RAG)~\cite{NEURIPS2020_6b493230}. In cybersecurity, command lines provide invaluable information for detecting malicious attacks by comparing them with known historical malicious command lines from a semantic perspective. However, the flexibility in command-line syntax and structure poses challenges for fully leveraging this information. For example, as shown in Fig.~\ref{fig:cover}, one can still correlate the two command lines according to their outputs despite the different appearances. To achieve this, using a robust embedding model to calculate the semantic similarity of command lines is promising. However, the grammatical differences between command lines and natural language sentences hinder the direct application of sentence embedding models to command-line tasks. Furthermore, one main challenge exacerbates the difficulty of research in command-line embedding: the scarcity of datasets specifically designed for command-line embedding tasks, both for training models and for fairly evaluating the performance of different methods.


To address the aforementioned challenges, this paper introduces the first comprehensive dataset, \dataset, which includes semantically similar pairs of command lines for both training and evaluating command-line embedding methodologies. Inspired by the successes of data synthesis by LLMs~\cite{wang-etal-2023-self-instruct, wang2023improvinge5_mistral}, the similar command-line pairs in our training set are automatically generated from a set of diverse command-line seeds initialized from multiple real-world sources by a total of six distinct LLMs trained on diverse datasets (\S~\ref{sec:training_data_collection}). This facilitates a broader range of command-line generation. For the testing set of \dataset, to prevent training data leakage, we directly employed a totally different data source instead of synthesizing command lines by LLMs, as done in the training set. (\S~\ref{sec:testing_data_collection})

To the end, our training set consists of 28,520 similar command-line pairs, totaling 55,909 unique command lines, and our testing set comprises 2,807 similar command-line pairs, totaling 5,576 unique command lines. Our dataset analysis and human evaluation results (\S~\ref{sec:eval_dataset}) demonstrate that our pipeline can generate highly diverse and high-quality similar command-line pairs.

Based on our proposed dataset, \dataset, we also developed the first embedding model specialized for command-line embeddings, called \sys. By encouraging semantically similar samples to come closer and simultaneously increasing the distance between semantically dissimilar samples in the embedding space, \sys can embed command lines into vectors from a semantic perspective. As demonstrated in Fig.~\ref{fig:cover}, even when command lines differ in appearance, \sys can still position them closely in the embedding space based on their semantic meanings.

Our evaluation results (\S\ref{sec:eval_model}) demonstrate that even the smallest version of \sys, with approximately 0.03 billion parameters, can surpass SOTA sentence embedding models with ten times more parameters (0.335 billion parameters) across various command-line specific tasks, such as malicious command-line detection, similar command-line retrieval, and command-line classification.

Our contribution is threefold. First, we propose the first dataset of similar command-line pairs named \dataset, which allows for training and performance evaluation. Through detailed validation of the dataset's effectiveness, we believe it is well-suited for further command-line research. Secondly, we explore the potential of using LLMs to synthesize command-line data in the cybersecurity domain. Our experiments demonstrate that LLMs can indeed generate high-quality and diverse data. Lastly, we propose the first semantic command-line embedding model, \sys. Our evaluations reveal that a command-line-specific embedding model significantly enhances performance across various downstream tasks compared to generic sentence embedding models. We open-source the entire dataset, model weights, and all program codes under BSD License at \href{https://github.com/cycraft-corp/CmdCaliper}{GitHub Repo}\footnote{\texttt{\scriptsize https://github.com/cycraft-corp/CmdCaliper}}

%% file: emnlp2023-latex/paper_content/related.tex
\section{Related Work}
\label{sec:related}

\subsection{Semantic Embedding}

Early works of sentence embedding such as Word2Vec~\cite{MCCD13} and Glove~\cite{pennington-etal-2014-glove}, require the training of a pre-defined static embedding lookup table to fuse into the embedding vector of different sentences.

Recent works leverage well-trained language models such as BERT~\cite{devlin-etal-2019-bert} and T5~\cite{t5} as pre-trained models to globally embed inputs (i.e., the words of sentences) into embedding vectors while considering contextual relationships to achieve impressive downstream performance. To fine-tune such models, a contrastive learning scheme~\cite{infonce} can be adopted, as demonstrated in \cite{gao2021simcse, chuang2022diffcse, zeng-etal-2022-contrastive, openai_embedding, wang2022e5, wang2023improvinge5_mistral}

In the cybersecurity domain, semantic embedding is crucial for computing semantic similarity across various data types, including logs and command lines. For log-based data, ~\citet{GE21} introduced an autoencoder-based model to convert log data into embedding vectors for anomaly detection. Log2Vec~\cite{LWZJXM19} creates a heterogeneous graph for each log dataset and uses the random walk method with Word2Vec to embed log data. LogBert~\cite{logbert} leverages BERT for anomaly detection, clustering the embedding vectors of normal samples to increase the separation from abnormal samples.

For command-line based data,~\citet{OSOTTNSOP21} adapted the tokenization methodology for command-line-based data and followed the Word2Vec approach to train a pre-defined embedding lookup table for a large amount of command-line data. Conversely, ~\citet{DWNSWLLX23} directly adopted Word2Vec for the command-line embedding.

\subsection{LLMs in Cybersecurity}
In the field of cybersecurity, many researchers~\cite{LLMinCybersecurity, LLMcyberopportunities} have also explored leveraging LLMs~\cite{gpt4turbo, claude3_haiku, llamav2} for malicious code generations.


~\citet{llms_script_kiddle} use LLMs to generate executable code for agent actions, facilitating automated cyber campaigns. ~\citet{LLMhoneypot} explore using LLMs as honeypots by generating executable commands to simulate Linux, Mac, and Windows terminals. ~\citet{llm_bypassing} demonstrate ChatGPT's ability to generate malicious code that can evade detection.

%% file: emnlp2023-latex/paper_content/approach_dataset.tex
\section{CmdDataset: The First Command-Line Similarity Dataset}
\label{sec:CmdDataset}

\begin{figure}[t]
\begin{center}
   \includegraphics[width=1\linewidth]{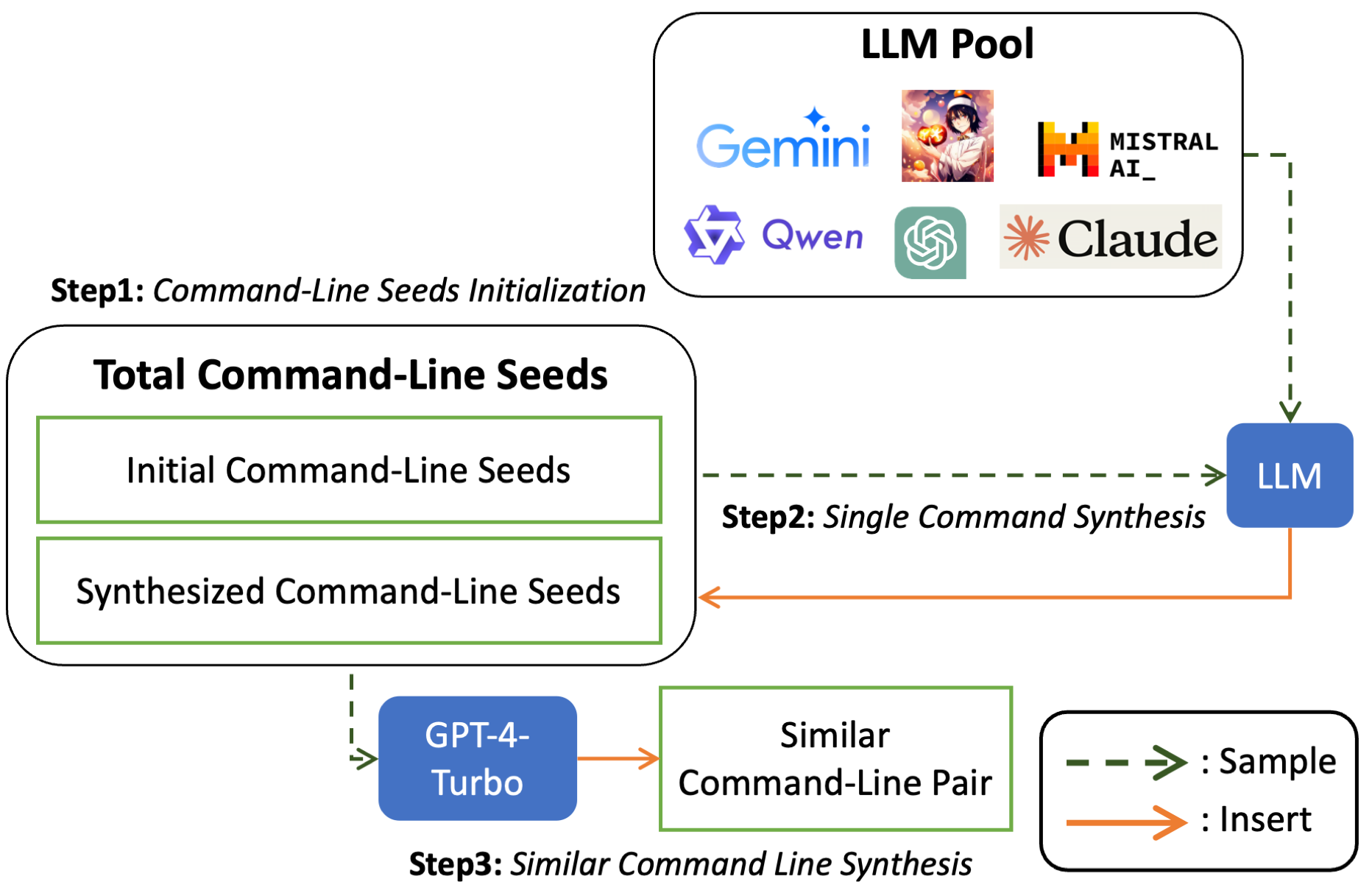}
\end{center}
   \caption{The illustration of the pipeline for automatically generating a dataset of similar command-line pairs using the Self-Instruct algorithm with a pool of LLMs.}
\label{fig:generation_pipeline}
\end{figure}

Despite the impressive performance of existing sentence embedding models~\cite{li2023towards, gao2021simcse, openai_embedding}, no embedding model has been designed specifically for command lines. We believe this is due to the unavailability of a large, diversified dataset with adequate annotations for effective training and unbiased evaluation.


In this section, we primarily focus on introducing the first command-line similarity dataset, named \dataset. This training set comprises 28,520 pairs of command lines automatically generated by a pool of LLMs. In contrast, the testing set contains 2,807 pairs of command lines collected from real-world attack scenarios.

\subsection{Training Set Synthesis by LLMs}
\label{sec:training_data_collection}



Collecting large-scale unlabeled or small-scale annotated command-line data is challenging due to two main factors. Firstly, labeling command-line datasets requires specialized cybersecurity knowledge, making it more stringent and costly than labeling images or natural language sentences. Secondly, privacy concerns involving company or personal information in real-world command lines discourage sharing, complicating efforts to gather diverse, large-scale datasets.


The automatic data generation process known as Self-Instruct \cite{honovich-etal-2023-unnatural, alpaca, wang-etal-2023-self-instruct} has proven effective in acquiring a comprehensive and diverse corpus of instructional data for fine-tuning LLMs. This process utilizes a powerful pre-trained large-scale language model, such as ChatGPT or Claude 3.

Our research is inspired by the substantial success of LLMs in code generation~\cite{codellama, patil2023gorilla} that exhibits similar structures to command lines and strong comprehension in the cybersecurity domain, such as malware generation~\cite{attacker_dream, gpthreats, llm_to_mitre, llm_bypassing}. Based on these capabilities, we adapt the Self-Instruct method to synthesize a substantial number of similar command-line pairs using LLMs. Our data synthesis pipeline comprises three stages: 1) Initial Seeds Collection, 2) Single Command Line Synthesis using a Pool of LLMs, and 3) Similar Command Line Synthesis, as illustrated in Fig.~\ref{fig:generation_pipeline}.

\subsubsection{Initial Seeds Collection}


Incorporating randomness into prompts is crucial for enabling LLMs to synthesize diverse command lines. This is achieved using initial seeds, which consist of a diverse set of command lines. During each synthesis iteration, a subset of these seeds is sampled to construct the prompt, diversifying it and encouraging a varied output from LLMs. To ensure high-quality initial seeds, we collected 2,061 diverse Windows command lines from multiple sources (e.g., public red-team exercises and Windows commands documentation). For more details on the initial-seed collection, see Appendix \ref{appendix:seed_collection}.

\subsubsection{Single Command Line Synthesis with a Pool of LLMs}
\label{sec:llm_pool}
Beyond the diversity of command-line seeds, the ability of LLMs to generate sufficiently varied data is also crucial. In the original Self-Instruct pipeline \cite{wang-etal-2023-self-instruct}, GPT-3~\cite{gpt3} was adopted for data generation, which confines all generated data to the distribution of GPT-3's training data. We extended this method by constructing a pool of distinct LLMs to aid in data generation. This is intuitive because the distribution of training data varies among different LLMs, leading to differences in the nature of the command lines they preferentially generate. The LLM pool of our pipeline comprises the following models: Mixtral 8x7B~\cite{mixtral8x7b}, WizardLM-13B-v1.2~\cite{xu2024wizardlm}, Gemini-1.0-Pro~\cite{gemini}, Claude-3-Haiku~\cite{claude3_haiku}, Qwen1.5-14B-Chat \cite{qwen}, and GPT-3.5-Turbo~\cite{chatgpt}. 

After constructing the LLM pool, we iteratively synthesize command lines by randomly sampling 12 command lines from the total command-line seeds--which include previously synthesized command lines and initial seeds--for prompt composition, as shown in Fig.~\ref{table:single_template}. We then instruct a randomly selected LLM from the pool to synthesize four new command lines distinct from those in the prompt. Valid command lines are extracted from the LLM responses. Due to the randomness of generation, the LLMs may not always produce four new valid command lines. These valid new command lines are added to the total command-line seeds for the next iteration. The generation process is halted after synthesizing 28,520 command lines.

In this step, LLMs may synthesize non-executable command lines with minor syntax errors. However, for our dataset of similar command-line pairs, ``similar'' refers to command lines sharing the same purposes or intentions, typically based on associated executable files, arguments, and argument values. Therefore, even with syntax errors, the command lines should still convey the same purpose or intention as their correct versions.

\subsubsection{Similar Command Line Synthesis}
\label{subsubsec:similar_cmd_synthesis}
After collecting 28,520 command lines, we instructed GPT-4-Turbo~\cite{gpt4turbo} to generate a similar command line for each. The prompting template for this instruction is displayed in Fig.~\ref{table:single_similar_template}. Here, ``similar'' refers to sharing the same purpose or intention, rather than merely having a similar appearance. This distinction is crucial, as in real-world scenarios, attackers may use different command lines or obfuscation techniques to achieve the same goal. We showcase several pairs of generated similar command lines in Table.~\ref{table:dataset_example}, demonstrating the efficacy of utilizing LLMs for this purpose.

\subsection{Real-World Testing Set Collection}
\label{sec:testing_data_collection}
To create a testing set that can fairly and comprehensively evaluate different methods, better reflect real-world usage scenarios, and avoid training data leakage, we neither directly partitioned the training set for the testing set, nor did we use LLMs to generate entirely new command lines from the initial seeds as the training set collection pipeline. Instead, we employed Splunk Attack data~\cite{splunk_attack_data}, a dataset curated from various attacks, as the source for our testing set. This allows us to evaluate various approaches in real-world scenarios, as it includes many malicious command lines corresponding to various MITRE ATT\&CK~\cite{miteattck} techniques, covering multiple distinct attack vectors.



Initially, we extracted 12,723 unique command lines from the Splunk Attack data. However, many command lines had similar meanings but differed slightly in appearance, leading to data duplication and potential evaluation inaccuracies. To address this concern, we generated explanations using ChatGPT and then converted these explanations into embeddings using GTE-Large~\cite{li2023towards}. We used these embeddings to remove command lines with semantically similar content, resulting in a final testing set of 2,807 command lines. For more details about the deduplication process, please refer to Appendix~\ref{appendix:deduplication_process}. We then followed the similar command line synthesis step proposed in Sec.~\ref{subsubsec:similar_cmd_synthesis} to instruct GPT-4-turbo to generate the corresponding similar command line for each command line.


For each command line from the original set of 2,807, we used the explanation embeddings to identify the 1,000 least similar command lines from the remaining 2,806 as negative command lines. Additionally, we designated the generated similar command line as the positive command line. This method avoids evaluation inaccuracies by preventing the inclusion of semantically similar command lines among the negatives.

%% file: emnlp2023-latex/paper_content/approach_model.tex
\section{\sys: A Semantic-Aware Command-Line Embedding Model}
\label{sec_cmdgpt}


Utilizing the proposed dataset \dataset, a command-line embedding model can be trained with a contrastive objective, like sentence embedding methodologies, as seen in \cite{gao2021simcse, chuang2022diffcse, openai_embedding}. Given an embedding model, denoted as $E$, the procedure of embedding a command line $c_i$ into an embedding vector $e_i$ can be described as $e_i = E(c_i)$. In each training iteration, several similar pairs are randomly sampled from the training set to form a batch, which contains $k$ similar command-line pairs $\{(x_i, x^+_i)\}^k_{i=1}$, where $(x_i, x^+_i)$ represents the $i^{\text{th}}$ similar command-line pair. Within the batch, the similar command line $x^+_i$ is regarded as a positive sample of sample $x_i$, thereby encouraging the associated embedding vectors to be closer within the feature space. Conversely, other samples $\{x^+_j, \forall j \in \{1, 2, \ldots, k\} \setminus \{i\}\}$ are treated as in-batch negatives \cite{in-batch-negative, openai_embedding, gao2021simcse}, encouraging the embedding vectors to be farther. 
The InfoNCE loss \cite{infonce}, denoted as $\mathcal{L}_{info}$, can be calculated as follows:
\begin{equation}
\label{eq:infonce_loss}
    \mathcal{L}_{info} = -\sum_{i=1}^k\log\frac{\exp(\frac{E(x_i)\cdot E(x^+_i)}{\tau})}{\sum\limits^k_{j=1}\exp(\frac{E(x_i)\cdot E(x^+_{j})}{\tau})},
\end{equation}
where $\tau$ is a hyperparameter that $\tau \in \mathbb{R}^+$.


%% file: emnlp2023-latex/paper_content/eval_dataset.tex
\section{Evaluation on \dataset}
\label{sec:eval_dataset}

\subsection{Experimental Settings}
In the entire \dataset synthesis pipeline, to enhance the diversity of the generated command lines, we follow the hyperparameter settings outlined in \cite{wang2023improvinge5_mistral}, setting the temperature parameter to 1 for all large language models (LLMs) to encourage more diverse outputs.

\subsection{The Statistics of the Dataset}

\begin{table}[]
\centering
\small

\begin{tabular}{c|c|c}
                                  & Training Set & Testing Set \\ \toprule
Num of command-line pair &   28,520 & 2,807  \\
Num of unique command line  &   55,909 & 5,576 \\
Max. command-line length   &    3,464  & 7,502 \\
Min. command-line length &  3  &  2   \\
Avg. command-line length & 91.635  &  96.301  \\
Std. of command-line length  &  60.794   &  196.675 
\\ \bottomrule
\end{tabular}%
\caption{The statistic information of \dataset.}
\label{table:statistic}
\end{table}

\begin{table}[]
\centering
\small

\begin{tabular}{c|c}
                                  & Coverage Rate (\%)\\ \toprule
Windows Commands &  73.52   \\
Windows Common File Extensions  &  70.67
\\ \bottomrule
\end{tabular}%
\caption{Coverage rates of \dataset across all Windows Commands and Windows common file extensions.}
\label{table:coverage}
\end{table}

The statistical information for the training and testing sets of the \dataset is presented in Table~\ref{table:statistic}. Thanks to the real-world sources of our testing set, the standard deviation of the testing data significantly differs from that of the training set, enabling a more generalized and accurate evaluation.

\subsection{The Diversity of the Synthesized Command Lines}
\label{sec:diversity_synthesis}



In this experiment, we aim to assess the diversity of the command lines synthesized by our data generation pipeline as described in Sec. \ref{sec:training_data_collection}. We excluded any initial command line seeds as well as similar paired command lines for this experiment. We calculated their coverage across all Windows commands\footnote{\href{https://learn.microsoft.com/en-us/windows-server/administration/windows-commands/windows-commands\#command-line-reference-a-z}{Windows Command-line reference A-Z}} and common file name extensions in a clean Windows 10 virtual machine, as demonstrated in Table~\ref{table:coverage}. To avoid bias, we excluded our manually formulated initial seeds and focused only on the command lines generated by LLMs. Overall, our synthesized command lines achieve a coverage rate of 73.52\% out of 306 unique Windows commands and 70.67\% out of 75 common file extensions. For more details about the coverage rate calculation process, please refer to Appendix~\ref{appendix:windows_coverage_rate}.

\begin{figure}[t]
\centering
\begin{center}
   \includegraphics[width=0.9\linewidth]{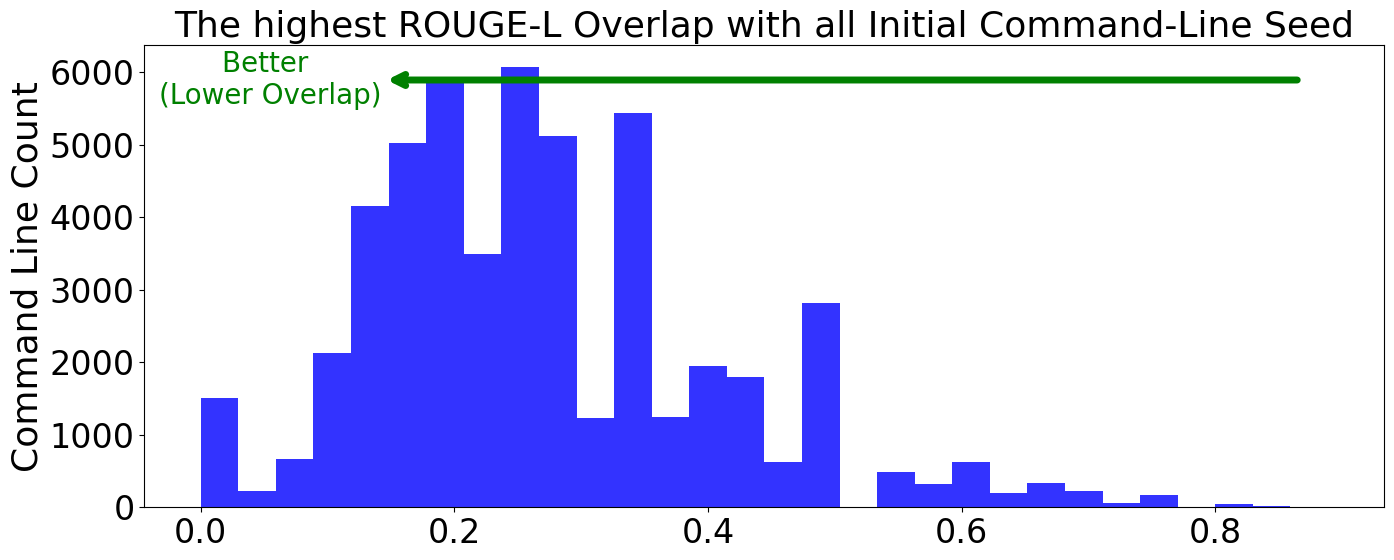}
\end{center}
   \caption{The distribution of the highest ROUGE-L overlap score  between the generated command lines and the initial command-line seeds.}
\label{fig:rouge}
\end{figure}

We conducted an in-depth analysis of the differences between the generated command lines and the initial command-line seeds, which served as a foundational starting point for constructing the command-line dataset. For each generated command line, we calculated the highest ROUGE-L overlap~\cite{lin-2004-rouge} which ranges from 0 to 1 among all initial command-line seeds. A higher ROUGE-L score indicates a greater overlap between the generated command lines and the initial command-line seeds. The distribution of ROUGE-L scores is illustrated in Fig.~\ref{fig:rouge}. These findings suggest that the command lines synthesized by our pipeline (Sec.~\ref{sec:training_data_collection}) are not limited to minor tweaks of the original command-line seeds. On the contrary, they are capable of producing a broad range of command lines, some of which may exhibit significant differences from the initial seeds.


\subsection{The Diversity within the Similar Command-Line Pairs}


\begin{figure}[t]
\centering
\begin{center}
   \includegraphics[width=0.9\linewidth]{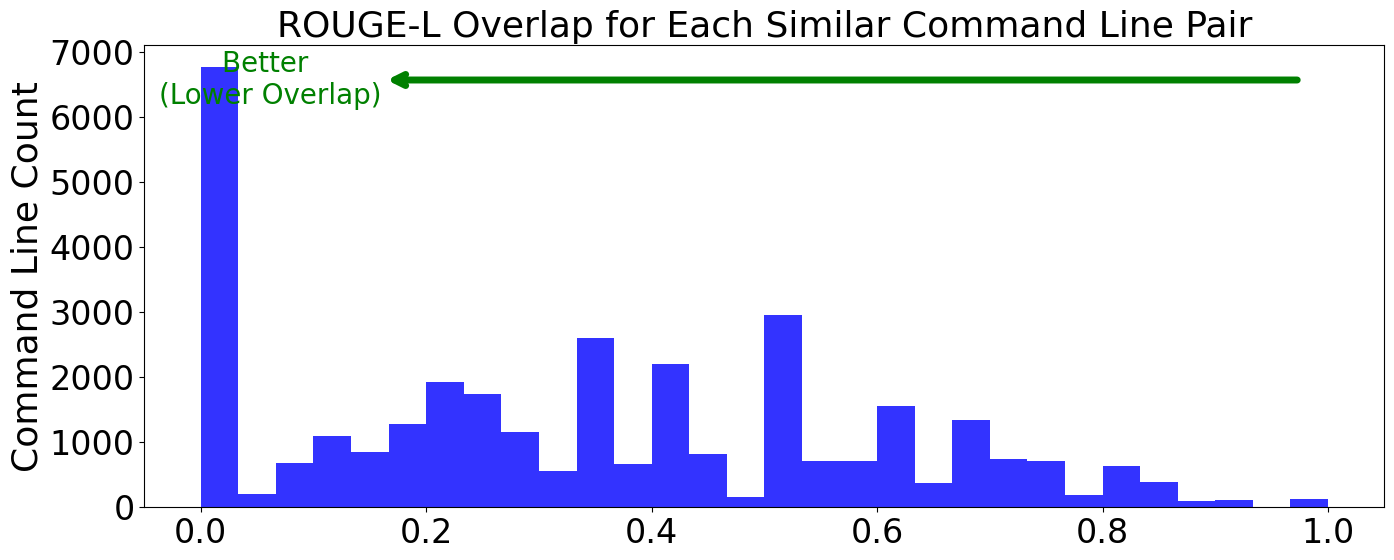}
\end{center}
   \caption{The distribution of ROUGE-L overlap score for all similar command-line pairs.}
\label{fig:pair_rouge}
\end{figure}

In this section, we examined the distribution of ROUGE-L overlap for each pair of similar command lines, as shown in Fig.~\ref{fig:pair_rouge}. These metrics help determine whether similar command-line pairs are derived from minor modifications to arguments or entirely different commands achieving a similar objective. Notably, our findings reveal that most ROUGE-L scores are low, nearing zero, suggesting that the ``similarity'' in our command-line dataset is not solely based on lexical similarities but rather reflects genuine semantic similarities. This vital understanding enables us to train command-line embedding models from a semantic perspective and subsequently evaluate the performance of different command-line embedding models.

\subsection{The Quality of the LLM's Command-Line Explanations}
\label{sec:explanation_quality}

In Sec.~\ref{sec:testing_data_collection}, we utilize ChatGPT~\cite{chatgpt} to generate explanations for command lines and employ GTE-Large~\cite{li2023towards} to convert these explanations into embeddings for data processing. In this evaluation, our focus is on assessing the quality of the explanations generated by LLM.


We randomly selected 200 command lines and their explanations from our training set. An expert (collaborator of this work) with over three years of cybersecurity experience assessed the correctness of each explanation, providing scores as positive, neutral, or negative, while ignoring minor syntax errors. We normalized the scores of the 200 evaluated command lines by assigning 0.5 points for positive labels, 0.25 points for neutral labels, and 0 points for negative labels, resulting in a total score of 100. After applying this scoring method, we obtained a final normalized score of \textbf{98.25}, indicating that the majority of command-line explanations accurately describe their intended purposes. Several good and bad examples are listed in Table~\ref{table:explantion_example}.

\subsection{The Quality of the Similar Command-Line Pairs}
\begin{figure}[t]
\begin{center}
   \includegraphics[width=0.9\linewidth]{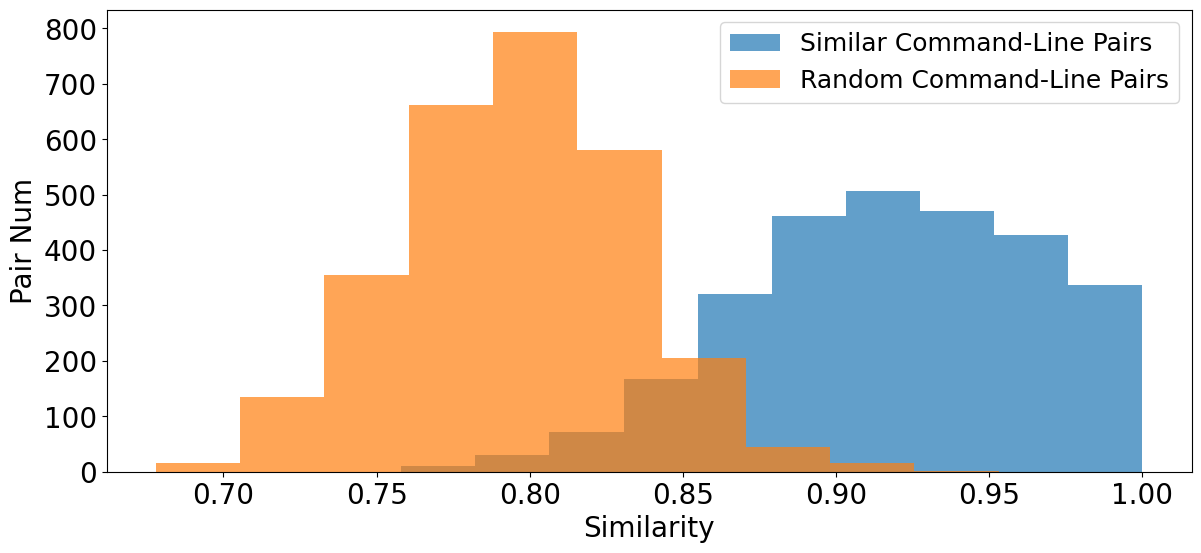}
\end{center}
   \caption{The histogram of the explanation similarity between random command-line pairs and similar command-line pairs in the testing set of \dataset.}
\label{fig:explanation_compare}
\end{figure}

In this experiment, we investigate whether the command-line pairs generated by LLMs are truly similar in terms of semantics. Leveraging the findings presented in Sec.~\ref{sec:explanation_quality}, which demonstrate a high alignment between the explanations generated by LLMs and those provided by human experts, we follow the same experimental settings, utilizing identical prompts and instructing ChatGPT~\cite{chatgpt} to generate explanations for each command line in the testing set.

After acquiring all explanations, we employ GTE-Large~\cite{li2023towards} to embed all explanations into corresponding embedding vectors and calculate the similarity between explanations for each similar command-line pair. Subsequently, we randomly construct an equal number of command-line pairs, totaling 2,807 pairs, with both command lines in each pair randomly sampled from all command lines in the testing set. 



Fig.~\ref{fig:explanation_compare} illustrates the similarity distributions of both similar and random command-line pairs. Notably, there is a significant gap in the explanation similarity between these two groups. Specifically, the average explanation similarity of each similar command-line pair exceeds that of 97.483\% of the random command-line pairs, indicating the effectiveness of synthesizing similar command lines with LLMs. Note that the range of similarity for GTE-Large is approximately between 0.65 and 1.




\subsection{Effectiveness of a Pool of LLMs}
\begin{table}[]
\small
\centering
\begin{tabular}{c|c}
    LLMs for Data Synthesis     & \begin{tabular}[c]{@{}c@{}}Coverage \\ Rate (\%)\end{tabular} \\ \toprule
GPT-3.5-Turbo~\cite{chatgpt}  &        48.75          \\
Mixtral 8x7B~\cite{mixtral8x7b}  &         27.5              \\
WizardLM-13B-v1.2~\cite{xu2024wizardlm}  &     35              \\
Gemini-1.0-Pro~\cite{gemini}  &        51.25              \\
Qwen1.5-14B-Chat~\cite{qwen} &        23.75        \\ 
Claude-3-Haiku~\cite{claude3_haiku}  &        30         \\ \midrule
LLM Pool (our) &        \textbf{70}   \\ \bottomrule
\end{tabular}%
\caption{Explanation clusters coverage rates of the command lines synthesized by different LLMs.}
\label{table:explanation_coverage_rate}
\end{table}

In this experiment, we aim to study whether a pool of LLMs can make the synthesized command lines more diverse versus utilizing a single LLM as we described in Sec.~\ref{sec:training_data_collection}. First, we instruct each LLM in our pool to synthesize 7,500 command lines from the same initial seeds, totaling 52,500 command lines. Using findings from Sec.\ref{sec:explanation_quality}, we then generate high-quality explanations for all synthesized command lines. Next, we embed these explanations into vectors using GTE-Large~\cite{li2023towards} and cluster them with DBSCAN~\cite{dbscan}, using a maximum distance of 0.08, a minimum of 5 samples per cluster, and cosine similarity as the distance metric. This process resulted in 80 distinct clusters (excluding the noise cluster), each representing a specific purpose or intention based on similar command line explanations. We then computed the coverage rates of the LLM pool and each individual LLM across these clusters to assess the diversity of synthesized command lines.

The results are presented in Table.~\ref{table:explanation_coverage_rate}. It is evident that command lines synthesized by the LLM pool cover the highest number of explanation clusters, reaching up to 70\%. This highlights the capability of utilizing a pool of LLMs pre-trained on diverse training data to generate a broader range of command lines.

%% file: emnlp2023-latex/paper_content/eval_model.tex
\section{Evaluation on \sys}
\label{sec:eval_model}
\subsection{Experimental Settings}
\sys was trained on three distinct model scales: small, base, and large, which are initialized from the GTE-small, -base, and -large~\cite{li2023towards}, respectively. For more details about the hyperparameters and training processes, please refer to to Appendix~\ref{appendix:cmdcaliper_hyperparameter}.

\subsection{Compare with SOTAs}
\begin{table}[]
\small
\centering
\begin{tabular}{c|clll}
Methods & \begin{tabular}[c]{@{}c@{}}MRR \\ @3\end{tabular} & \begin{tabular}[c]{@{}c@{}}MRR \\ @10\end{tabular} & \begin{tabular}[c]{@{}c@{}}Top \\ @3\end{tabular} & \begin{tabular}[c]{@{}c@{}}Top \\ @10\end{tabular} \\ \toprule
\begin{tabular}[c]{@{}c@{}}Levenshtein \\ distance\textsuperscript{1} \end{tabular}   &   71.23   &   72.45    &   74.99    &    81.83   \\ 
Word2Vec\textsuperscript{2}  &  45.83    &  46.93    &  48.49   &  54.86   \\ \midrule
E5$_{S}$\textsuperscript{3}  & 81.59 & 82.6 & 84.97 & 90.59 \\ 
GTE$_{S}$\textsuperscript{4}  & 82.35 & 83.28 & 85.39 & 90.84 \\ 
CmdCaliper$_{S}$       & \textbf{86.81} & \textbf{87.78} & \textbf{89.21} & \textbf{94.76}  \\ \midrule

BGE-en$_{B}$\textsuperscript{5}  & 79.49 & 80.41 & 82.33 & 87.39 \\ 
E5$_{B}$    & 83.16 & 84.07 & 86.14 & 91.56 \\ 
GTR$_{B}$\textsuperscript{6}   & 81.55 & 82.51 & 84.54 & 90.1 \\ 
GTE$_{B}$    & 78.2 & 79.07 & 81.22 & 86.14 \\ 
CmdCaliper$_{B}$    & \textbf{87.56} & \textbf{88.47} & \textbf{90.27} & \textbf{95.26} \\ \midrule

BGE-en$_{L}$   & 84.11 & 84.92 & 86.64 & 91.09 \\ 
E5$_{L}$   & 84.12 & 85.04 & 87.32 & 92.59 \\ 
GTR$_{L}$  & 88.09 & 88.68 & 91.27 & 94.58 \\ 
GTE$_{L}$  & 84.26 & 85.03 & 87.14 & 91.41 \\
CmdCaliper$_{L}$ & \textbf{89.12} & \textbf{89.91} & \textbf{91.45} & \textbf{95.65} \\ \bottomrule
\end{tabular}%

\textsuperscript{1}\cite{le_distance}
\textsuperscript{2}\cite{MCCD13}
\textsuperscript{3}\cite{wang2022e5}
\textsuperscript{4}\cite{li2023towards}
\textsuperscript{5}\cite{bge_embedding}
\textsuperscript{6}\cite{ni-etal-2022-large_gtr}

\caption{Comparison with the SOTAs for different pre-trained language models. Subscript $S$, $B$, and $L$ denote the Small (0.03 billion parameters), Base (0.11 billion parameters), and Large (0.34 billion parameters) versions, respectively.}
\label{table:cmdcaliper_comparison}
\end{table}

This section compares several SOTA sentence embedding methods using the testing set from \dataset. We adopt Mean Reciprocal Ranking@K (MRR) and Top@K metrics for evaluating the performance of \sys, following the text search task methodology~\cite{muennighoff2022mteb}. For more details about the two evaluation metrics, please refer to Appendix~\ref{appendix:evaluation_metrics}. Both metrics yield scores between 0 and 100. The results of this comparative analysis are presented in Table~\ref{table:cmdcaliper_comparison}.

The results indicate that \sys consistently achieves competitive performance with state-of-the-art (SOTA) models across all evaluation metrics and scales. Specifically, \sys-Base achieved an MRR@3 score of 87.56, surpassing the embedding model we fine-tuned on - GTE-Base~\cite{li2023towards} by 9.36 and outperforming all sentence embedding models of comparable size. On a larger scale, \sys-Large achieved an MRR@3 score of 89.12, surpassing GTE-Large by 4.86. Remarkably, even \sys-Small, with a mere 0.03B parameters, is comparable with all SOTA embedding models at the large scale (with 0.335B parameters).



\subsection{Semantic-Based Malicious Command-Line Detection}
\begin{table}[]
\small
\centering
\begin{tabular}{c|cccc}
\toprule
Models \textbf{\textbackslash} $r$ (\%) & 20  & 40  & 60  & 80  \\ \midrule
GTR$_{Base}$                                 & 0.793 & 0.852 & 0.866 & 0.903 \\ 
E5$_{Base}$                                    & 0.796 & 0.859 & 0.87  & 0.899 \\ 
GTE$_{Base}$                               & 0.8   & 0.868 & 0.874 & 0.903 \\
CmdCaliper$_{Base}$                          & \textbf{0.869} & \textbf{0.906} & \textbf{0.927} & \textbf{0.939} \\ \bottomrule
\end{tabular}%
\caption{The AUC comparison for different embedding models and different sample rate $r\%$.}

\label{table:malicious_detection}
\end{table}


In this section, we approach malicious command-line detection as a retrieval task using the open-source atomic-red-team dataset~\cite{atomic_red_team}, which includes command lines corresponding to 55 different MITRE ATT\&CK techniques~\cite{miteattck} (i.e., each technique describes different command line attack behaviors). We iteratively select $r\%$ of the malicious command lines from each technique as query command lines, while the remaining $(1 - r)\%$ serve as positive command lines. Command lines from other techniques act as negative command lines. This process is repeated for each technique, and we calculate the average area under curve (AUC). The intuition behind this experiment is that a good embedding model should cluster malicious command lines from the same technique closer together, as they share similar attack behaviors. For more details about the experiment setup, please refer to Appendix~\ref{appendix:semantical_malicious_detection}.

The detection results are illustrated in Table.~\ref{table:malicious_detection}. As observed, CmdCaliper-Base significantly outperforms all embedding models not fine-tuned on the command-line dataset. This difference is especially pronounced when the sample ratio is smaller. For instance, at a 20\% sample ratio ($r=20$), CmdCaliper-Base improves upon GTE-Base~\cite{li2023towards} by approximately 0.069 in AUC. This suggests that when the query command-line set is smaller, the model requires a deeper understanding of the semantics of command lines.

\subsection{Transfer to Command-Line Classification}

\begin{table}[]
\centering
\small
\begin{tabular}{c|c|c}
\toprule
Embedding Models & \begin{tabular}[c]{@{}c@{}}Params \\ (B)\end{tabular}  & Accuracy (\%)  \\ \midrule
E5$_{Base}$  & 0.11 &  93.86  \\
GTE$_{Base}$ & 0.11 & 92.8 \\
CmdCaliper$_{Base}$ & 0.11 & \textbf{96.37} \\\bottomrule
\end{tabular}%
\caption{Accuracy comparison for command-line classification fine-tuned on fixed embedding models.}
\label{table:transfer}
\end{table}

Fine-tuning an additional module on a pre-trained embedding model for tasks like classification or regression often outperforms training from scratch. This is because well-trained embeddings capture rich, meaningful information that can be used across tasks. In this experiment, we trained a logistic regression classifier for Windows command classification using fixed command-line embeddings from different approaches.

Collecting a labeled command-line dataset poses significant challenges. To address this, we selected seven Windows commands: `find', `robocopy', `msiexec', `rundll32', `sc query', `certutil', and `print' to synthesize 24,500 training and 24,500 testing command lines. Each command class contains an equal number of examples, ensuring a balanced dataset for fair evaluation. For more details about the the classification dataset synthesis and the experimental setup, please refer to Appendix~\ref{appendix:classification_dataset_synthesis}.

The results of the classification are presented in Table.~\ref{table:transfer}. As observed, CmdCaliper generally outperformed other sentence embedding models in terms of accuracy for the same model size. For example, CmdCaliper-Base achieved a 3.57\% improvement over GTE-Base~\cite{li2023towards}. These findings highlight the importance of specialized embedding models for command-line data, allowing the models to encode more command-line information into their embedding vectors.

\subsection{Does Command-Line Embedding Benefit from Sentence Embedding?}
\begin{table}[]
\centering
\small
\begin{tabular}{c|c|clll}
\toprule
Model & \begin{tabular}[c]{@{}c@{}}Params \\ (B)\end{tabular}  & \begin{tabular}[c]{@{}c@{}}MRR \\ @3\end{tabular} & \begin{tabular}[c]{@{}c@{}}MRR \\ @10\end{tabular} & \begin{tabular}[c]{@{}c@{}}Top \\ @3\end{tabular} & \begin{tabular}[c]{@{}c@{}}Top \\ @10\end{tabular} \\ \midrule
\begin{tabular}[c]{@{}c@{}}Random \\ Initialization \end{tabular} & 0.11 &  70.43 & 72.14 & 74.49 & 84.04 \\
Bert$_{Base}$ & 0.11  & 82.25 & 83.38 & 85.79 & 92.13 \\
GTE$_{Base}$ & 0.11 & \textbf{87.56} & \textbf{88.47} & \textbf{90.27} & \textbf{95.26} \\ \bottomrule
\end{tabular}%
\caption{Comparison of the performance of CmdCaliper fine-tuning from the different model.}
\label{table:benefit_sentence_embedding}
\end{table}





We conducted experiments under three settings: training the Bert-Base~\cite{devlin-etal-2019-bert} network architecture, the pretrained model of GTE-Base~\cite{li2023towards}, with randomly initialized weights; fine-tuning from Bert-Base; and fine-tuning directly from the GTE-Base model. The results of the comparison are illustrated in Table~\ref{table:benefit_sentence_embedding}. As observed, the performance of the pretrained model significantly influences the performance of the command-line embedding model. For instance, fine-tuning from the embedding model yields the highest MRR@3, showing a 5.31\% improvement compared to direct fine-tuning with BERT.

We believe that the reason command-line embedding models benefit from a good sentence embedding model lies in the fact that, although command lines often have entirely different grammar and structure from general sentences, in many cases, we can still infer some partial meanings of the command lines from semantically meaningful words such as filenames, arguments, or folder names.

%% file: emnlp2023-latex/paper_content/conclusion.tex
\section{Conclusion}
\label{sec:conclusion}


In this work, we introduce \dataset, a dataset of similar command-line pairs. The training set utilizes the impressive capabilities of an LLM pool for automated generation, while the testing set consists of real-world malicious command lines for realistic evaluation. We also present \sys, the first dedicated command-line embedding model. Our results show that \sys, specifically designed for command-line processing, outperforms existing sentence embedding methods in various command-line downstream tasks, such as command classification, malicious command line detection, and similar command-line retrieval.

We open-source the dataset, model weights, and all program codes, hoping this study sheds light on future command-line embedding research.

%% file: emnlp2023-latex/paper_content/limitation.tex
\section{Limitation}
Despite the contributions made in this paper, several tasks listed below are worth exploring in the future.
\begin{itemize} 
  \item Support more command-line interpreters: Given their prominence in recent cybersecurity incidents, this study focuses on Windows and PowerShell commands. Compelling statistics~\cite{mtrends2023} reveal that over two-thirds of script-based attacks leverage PowerShell. Nevertheless, an investigation into a unified command-line embedding model capable of spanning multiple command-line interpreters presents a promising avenue for future research.
  \item Resilience against command-line obfuscation: While \sys was trained on semantically similar command-line pairs, providing a certain degree of resilience against obfuscated command lines in this study, attackers often employ more sophisticated command-line obfuscation techniques to evade defense and detection mechanisms. This poses a significant challenge when relying solely on the embedding model for detecting malicious activities.  
  \item Nested command line: We can obtain the corresponding semantic embedding vector by inputting a command line into \sys. However, a command line itself can be a composition of multiple command lines as well, making it difficult to accurately embed them into the feature space. Techniques such as few-shot learning~\cite{gpt3} or instruction-finetuned text embeddings~\cite{su-etal-2023-one} may provide potential solutions for generating command-line embeddings for specific downstream tasks. This area represents another possibility for future investigation.
\end{itemize}

%% file: emnlp2023-latex/paper_content/acknowledgment.tex
\section{Acknowledgment}
We would like to express our deepest gratitude to the renowned cybersecurity expert Ming-Chang Chiu (aka Birdman) for providing invaluable insights and expertise. We also thank Chi-Shun Chen for sharing his experience in malicious command-line hunting within large critical infrastructure in the early stage of this research.

%% file: emnlp2023-latex/paper_content/appendix.tex
\section{Testing Set Collection Detail}
\label{appendix:deduplication_process}
To deduplicate, we utilized ChatGPT~\cite{chatgpt} to transform all command lines into concise descriptions that encapsulate the purpose and intention. Subsequently, we transformed these brief descriptions into embeddings using GTE-Large~\cite{li2023towards}, which achieved SOTA performance on the MTEB leaderboard~\cite{muennighoff2022mteb} among models of similar size. We then applied DBSCAN~\cite{dbscan} for clustering the embeddings. Through this approach, each cluster contains command lines with highly similar semantics based on their explanations. Finally, we extracted two command lines from each cluster, resulting in a testing set comprising 2,807 command lines in total.

\section{Initial-Seed Collection Detail}
\label{appendix:seed_collection}
To gather high-quality initial seeds, we first extracted all command lines executed in DARPA Transparent Computing~\cite{darpa}, totaling 142,886 unique command lines. We then applied a heuristic filtering process to eliminate command lines that are semantically similar and differ only slightly, such as variations in log file suffixes. Ultimately, we curated 722 command lines as part of the initial seeds.

To further extend the initial seeds, we formulated 796 command lines based on the descriptions and corresponding syntax found in Windows Commands\footnote{\href{https://learn.microsoft.com/en-us/windows-server/administration/windows-commands/windows-commands\#command-line-reference-a-z}{Windows Command-line reference A-Z}}. Additionally, we parsed all example command lines from SS64\footnote{\href{https://ss64.com/nt/commands.html}{SS64 Windows CMDs}}, totaling 497 command lines, and collected an additional 46 command lines from GitHub. Together, these contributions amounted to 2,061 high-quality and diverse command lines, which were integrated to form our initial command-line seeds.

\begin{table*}[]
\begin{tabularx}{\linewidth}{|X|X|}
\toprule

\begin{minipage}{\dimexpr\linewidth-2\fboxsep-\fboxrule}
\vspace{1mm}
\tt\raggedright
- "C:\textbackslash{}Windows\textbackslash{}System32\textbackslash{}bitsadmin.exe" /transfer 59697582645 /priority foreground  http://example.com/example1234 "C:\textbackslash{}Users\textbackslash{}Public\textbackslash{}Videos\textbackslash{}V123456789\textbackslash{}log32.dll"\\
- powershell -command "Start-BitsTransfer -Source 'http://malicious.com/malicious1234' -Destination 'C:\textbackslash{}Users\textbackslash{}Public\textbackslash{}Videos\textbackslash{}V99999999\textbackslash{}log32.dll' -Priority High"
\vspace{1mm}
\end{minipage} \\ \midrule

\begin{minipage}{\dimexpr\linewidth-2\fboxsep-2\fboxrule}
\vspace{1mm}\tt\raggedright
- "cmd" /c "net use \textbackslash{}\textbackslash{}REMOTEDIR /user:Administrator password /persistent:no"\\
- python -c "import os; os.system('net use \textbackslash{}\textbackslash{}REMOTEDIR /user:Administrator password /persistent:no')"
\vspace{1mm}
\end{minipage} \\ \midrule

\begin{minipage}{\dimexpr\linewidth-2\fboxsep-2\fboxrule}
\vspace{1mm}\tt\raggedright
- reg query "HKLM\textbackslash{}Software\textbackslash{}Microsoft\textbackslash{}Windows\textbackslash{}CurrentVersion\textbackslash{}Uninstall" /f "Chrome" /s\\ - Get-ItemProperty HKLM:\textbackslash{}Software\textbackslash{}Microsoft\textbackslash{}Windows\textbackslash{}CurrentVersion\textbackslash{}Uninstall\textbackslash{}* | Select-Object -Property DisplayName, UninstallString | Where-Object \{\$\_.DisplayName -like '*Chrome*'\} | Format-Table -AutoSize 
\vspace{1mm}
\end{minipage} \\ \midrule

\begin{minipage}{\dimexpr\linewidth-2\fboxsep-2\fboxrule}
\vspace{1mm}\tt\raggedright
- schtasks /create /tn "TaskName" /tr "C:\textbackslash{}Path\textbackslash{}to\textbackslash{}program.exe" /sc daily /st 00:00\\ - cronjob schedule daily 00:00 /path/to/program 
\vspace{1mm}
\end{minipage} \\
\bottomrule
\end{tabularx}%
\caption{The similar command-line pairs in \dataset. Similar command lines are not merely similar on the lexical level but also in terms of their intrinsic purpose and semantic meaning.}
\label{table:dataset_example}
\end{table*}


\begin{figure*}[]
    \centering
    \normalsize
    \noindent\fbox{%
    \begin{minipage}{\dimexpr\linewidth-2\fboxsep-2\fboxrule}
\tt

Here are 12 Windows command line examples for referencing:\\
1. \textbf{\{sampled command line seed 1\}} \\
2. \textbf{\{sampled command line seed 2\}} \\
... \\
12. \textbf{\{sampled command line seed 12\}} \\
\\
Your job is to synthesize 4 new Windows command lines. Please adhere to the following synthesizing guidelines: \\
- Ensure diverse command lines in appearance, argument value, purpose, result, and length, particularly making sure the generated command lines differ significantly from the reference command lines in every aspect. \\
- Prioritize practicality in generated commands, ideally those executed or executable. For example, please give me real argument value, filename, IP address, and username. \\
- Include Windows native commands, commands from installed applications or packages (for entertainment, work, artistic, or daily purposes), commands usually adopted by IT, commands corresponding with mitre att\&ck techniques, or even some commonly used attack command lines. The more uncommon the command line, the better. \\
- Do not always generate short command lines only. Be creative to synthesize all kind of command lines. \\

Give me your generated command lines only without any explanation or anything else. Separate each generated command line with "\textbackslash n" and add a prefix "<CMD>" before each generated cmd.
\end{minipage}
}
\caption{The prompt used for generating a single command line. 12 exemplary command lines are randomly sampled from total command-line seeds for in-context demonstration.}
\label{table:single_template}
\end{figure*}

\begin{figure*}[]
    \centering
    \normalsize
    \noindent\fbox{%
    \begin{minipage}{\dimexpr\linewidth-2\fboxsep-2\fboxrule}
\tt
Your task is to generate a similar Windows command line for each entry in the following command line list. \\
In this task, 'similar' means that the command lines share the same purpose, or intention, rather than merely having a similar appearance. \\

Consequently, the generated command lines may differ significantly in argument values, format, and order from the original command line, or even from a different executable file, as long as they serve a similar purpose or intention. \\
\textbf{\{query command line\}} \\\\
Be creative to make the command lines appear distinctly different while adhering to the defined 'similar' criteria. 
For instance, you might employ obfuscation techniques, randomly rearrange the order of arguments, change the way to call the exe file, or substitute the executable file with a similar one. \\
Please provide only the generated similar command lines without any explanation, prefixed with "<CMD>", and separate each command line with "\textbackslash n".
\end{minipage}
}
\caption{The prompt used for generating similar command lines. The query command line is randomly sampled from the total command-line seeds.}
\label{table:single_similar_template}
\end{figure*}

\begin{table*}[]
\begin{tabularx}{\linewidth}{c|l|c}

Command Line & Explanation & Labels\\ \toprule

\begin{minipage}{\dimexpr0.45\linewidth-2\fboxsep-\fboxrule}
\vspace{1mm}
\tt\raggedright
net use Z: \textbackslash\textbackslash192.168.1.1\textbackslash{}SharedFolder /user:administrator Passw0rd! | findstr /i connected
\vspace{1mm}
\end{minipage} & 
\begin{minipage}{\dimexpr0.4\linewidth-2\fboxsep-\fboxrule}
\vspace{1mm}
\raggedright
This command line is used to map a network drive to the letter Z, connect to a shared folder on a specific IP address using administrator credentials, and then search for the keyword "connected" in the output.
\vspace{1mm}
\end{minipage} & Positive \\ \midrule

\begin{minipage}{\dimexpr0.45\linewidth-2\fboxsep-\fboxrule}
\vspace{1mm}
\tt\raggedright
schtasks /create /sc weekly /d MON,TUE,WED,THU,FRI /tn "WeeklyBackup" /tr "C:\textbackslash{}Scripts\textbackslash{}backup.bat" /st 18:00
\vspace{1mm}
\end{minipage} & 
\begin{minipage}{\dimexpr0.4\linewidth-2\fboxsep-\fboxrule}
\vspace{1mm}
\raggedright
This command line creates a scheduled task to run a backup script every weekday at 6:00 PM.
\vspace{1mm}
\end{minipage} & Positive \\ \midrule

\begin{minipage}{\dimexpr0.45\linewidth-2\fboxsep-\fboxrule}
\vspace{1mm}
\tt\raggedright
tasklist /fi "IMAGENAME eq notepad.exe" /fo list | find "1234"
\vspace{1mm}
\end{minipage} & 
\begin{minipage}{\dimexpr0.4\linewidth-2\fboxsep-\fboxrule}
\vspace{1mm}
\raggedright
This command line is used to list all running processes with the name "notepad.exe" and then search for a specific process ID "1234" within the list.
\vspace{1mm}
\end{minipage} & Positive \\ \midrule

\begin{minipage}{\dimexpr0.45\linewidth-2\fboxsep-\fboxrule}
\vspace{1mm}
\tt\raggedright
findstr /s /i /m "hello world" "world take care" C:\textbackslash{}Users\textbackslash{}* \textbackslash{}*.pdf
\vspace{1mm}
\end{minipage} & 
\begin{minipage}{\dimexpr0.4\linewidth-2\fboxsep-\fboxrule}
\vspace{1mm}
\raggedright
Search for the phrase "hello world" in all PDF files located in the C:\textbackslash{}Users directory and its subdirectories, ignoring case and only displaying the file names that contain the phrase.
\vspace{1mm}
\end{minipage} & Neutral \\ \bottomrule

\end{tabularx}%
\caption{Example of command lines and their corresponding explanations generated by GPT-3.5-Turbo~\cite{chatgpt}. The rightmost column denotes the labels (e.g., Positive, Neutral, or Negative) assigned by the expert.}
\label{table:explantion_example}
\end{table*}

\section{Windows-Commands Coverage Rate Detail}
\label{appendix:windows_coverage_rate}
While many Windows commands are listed separately, several utilize identical executable files, like `reg add' and `reg copy'. To present a unified perspective, commands with shared executable were further grouped into one, resulting in 306 unique Windows commands.

For common file name extensions, we first parsed all extensions from a clean Windows 10 virtual machine, and removing those with special characters and frequencies lower than 0.05\%. This process yielded a total of 75 common extensions. We then identified how many of these extensions are included in our training set.

\section{Evaluation Metrics Detail}
\label{appendix:evaluation_metrics}
MRR@K is a key metric in information retrieval and recommendation systems. It calculates the average reciprocal rank of the first relevant item within a list of ranked results, focusing on the top K positions. This method helps us gauge how well the most relevant item ranks among the top 10 with the highest predicted scores. The Top@K metric is similar to MRR@K but differs in that it awards a score if the ground truth is within the top K ranks, without the rank-dependent decay seen in MRR.

\section{Hyperparameters and Training Process of \sys Detail}
\label{appendix:cmdcaliper_hyperparameter}
We trained \sys for 2 epochs using the Adam optimizer~\cite{adam} with a learning rate of 0.00002 and a batch size of 64. For the temperature parameter $\tau$ in Equation~\ref{eq:infonce_loss}, we set it to 0.05. \sys was trained on three distinct model scales: small, base, and large. These models were initialized from the GTE-small~\cite{li2023towards}, GTE-base, and GTE-large, respectively.

We randomly selected 1,000 similar pairs from the training set to form a validation set. Evaluations on the validation set were conducted every 50 training steps. The checkpoint that demonstrated optimal performance on the validation set was then used for subsequent evaluations on the testing set.

\section{Semantic-Based Malicious Command-Line Detection Detail}
\label{appendix:semantical_malicious_detection}
Initially, we define the pre-collected set of malicious command lines as the `malicious gene pool'. Given a new command line, we leverage a command-line embedding model to obtain their embedding vectors, and compute the semantic similarity between each command line within the malicious gene pool. If one of the similarities exceeds a pre-defined threshold, we classify the new command line as malicious. This approach enables us to detect malicious command lines from a semantic perspective, even when attackers attempt to obfuscate command lines to evade pattern-based detection.

In this experiment, we utilize the open-source dataset: atomic-red-team~\cite{atomic_red_team}, which encompasses a variety of command lines corresponding to numerous MITRE ATT\&CK techniques (e.g., Abuse Elevation Control Mechanism or Browser Session Hijacking). The dataset consists of a set of techniques, denoted as $\{ t_1, t_2, \ldots, t_n \}$. Within each technique $t_i$, we obtain a set of malicious command lines, denoted as $\mathbb{L}_i$, containing $M_i$ entries: $\mathbb{L}_i=\{c^i_1, c^i_2, \ldots, c^i_{M_i}\}$. By unioning all these malicious command line sets, we form a total command line set $\mathbb{A}$, which consists of 1,523 malicious command lines across various techniques, represented as $\mathbb{A} = \bigcup_{i=1}^{n} \mathbb{L}_i$. To construct the malicious gene pool $\mathbb{P}_i$ for each technique $t_i$, we select the first $r\%$ of the command lines from each technique, defined as $\mathbb{P}_i = \{c^i_1, c^i_2, \ldots, c^i_{\left\lceil \frac{r}{100} \times M_i \right\rceil}\}$. The remaining command lines form the incoming command line set $\mathbb{O}_i$, denoted as: $\mathbb{O}_i = \{c^i_{\left\lceil \frac{r}{100} \times M_i \right\rceil+1}, \ldots, c^i_{M_i}\}$.

For malicious gene pool construction, we exclude techniques with fewer than 9 malicious command lines, resulting in a total of 55 distinct malicious gene pools corresponding to different techniques. For the evaluation of each technique $t_i$, we first exclude the command lines in the malicious gene pool to form a candidate command line set $\mathbb{C}_i$, denoted as $\mathbb{C}_i = \mathbb{A} \setminus \mathbb{P}_i$. We treat the incoming command line set $\mathbb{O}_i$ as the positive set, while the negative set $\mathbb{G}_i$ is formed by excluding all command lines corresponding to the technique $t_i$, denoted as $\mathbb{G}_i = \mathbb{A} \setminus \mathbb{L}_i$.

Given a command line $c^i_j$ from the candidate command line set $\mathbb{C}_i$ and an embedding model $E$, the embedding vector $e^i_j$ can be computed by $e^i_j = E(c^i_j)$. Subsequently, the malicious score $s_{c^i_j}$ of the command line $c^i_j$ for the technique $t_i$ is determined by calculating its maximum similarity with each command line $p^i_k$ in the malicious gene pool $\mathbb{P}_i$:
\begin{equation}
    s^i_{c^i_j} = \max_{p^i_k \in \mathbb{P}_i}S(e^i_j, E(p^i_k))
\end{equation}
where $S(\cdot)$ is the similarity function (e.g., cosine similarity function).
Note that if the command line $c_j$ belongs to the positive set $\mathbb{O}_i$ of the technique $t_i$, then the malicious score should exceed the pre-defined threshold $\tau$ for correct classification; otherwise, the score should be below the threshold if the command line is in the negative set $\mathbb{G}_i$.

We iteratively evaluated all 55 techniques and concatenated all malicious scores to compute the area under the curve (AUC). This evaluation metric aligns with real-world application scenarios where a static threshold is usually applied across all techniques. 

\section{Classification Dataset Synthesis Detail}
\label{appendix:classification_dataset_synthesis}
These commands to synthesize the command-line classification dataset were chosen based on their ability to accept a wide range of randomly generated strings as arguments, provided that the corresponding file exists. The classifier's task is to identify the corresponding Windows command, regardless of the argument's length or complexity. For example, the command lines ``find `fewj2po3kdlewfmpemrgborktig fe34krop4k5ogjs9rkgewfefw34f'" and ``find `test'" should both be correctly categorized under the `find' command. This highlights the importance of a robust command-line embedding model in encoding the command information within its embeddings, as it plays a crucial role in determining the purpose of each command line.

In this experiment, we used the pattern ``<command> `<argument value>'" to randomly generate 7,000 command lines for each command. Of these, 3,500 were assigned to the training set and the remaining 3,500 to the testing set. The arguments for each command line were formed by concatenating seven random strings, made up of ASCII letters and digits, with lengths ranging from 1 to 20 characters, separated by spaces. To increase the difficulty of classification and simulate the obfuscation techniques that attackers might use in real-world scenarios to evade detection, we also randomly incorporated seven different commands into the argument values. For example, a synthesized command line might read: ``certutil.exe `msiexec tr9QI1L find C print oGod 5K 7Okf4 2ZcVT9 rundll32 sc query mNjIL robocopy q5'", where only the first command, `certutil', is valid.

We randomly selected 20\% of the training set to serve as a validation set in the search for optimal hyperparameters. Subsequently, we utilized the obtained optimal hyperparameters to train a logistic regression classifier for performance evaluation on the testing set.